\pdfoutput=1
\documentclass[11pt]{article}

\usepackage[final]{coling}

\usepackage{times}
\usepackage{latexsym}

\usepackage[T1]{fontenc}
\usepackage[utf8]{inputenc}

\usepackage{microtype}

\usepackage{inconsolata}

\usepackage{graphicx}

\usepackage{amssymb}
\title{Bilingual BSARD: Extending Statutory Article Retrieval to Dutch}

\author{Ehsan Lotfi\thanks{Indicates equal contribution} \And Nikolay Banar\footnotemark[1]  \And Nerses Yuzbashyan \And Walter Daelemans \\ \AND
        CLiPS, University of Antwerp, Belgium \\ \texttt{\{ehsan.lotfi, nicolae.banari, nerses.yuzbashyan, walter.daelemans\}} \\\texttt{@uantwerpen.be}}

\begin{document}
\maketitle
\begin{abstract}
Statutory article retrieval plays a crucial role in making legal information more accessible to both laypeople and legal professionals. Multilingual countries like Belgium present unique challenges for retrieval models due to the need for handling legal issues in multiple languages. Building on the Belgian Statutory Article Retrieval Dataset (BSARD, \citet{louis2022statutory}) in French, we introduce the bilingual version of this dataset, bBSARD. The dataset contains parallel Belgian statutory articles in both French and Dutch, along with legal questions from BSARD and their Dutch translation. Using bBSARD, we conduct extensive benchmarking of retrieval models available for Dutch and French. Our benchmarking setup includes lexical models, zero-shot dense models, and fine-tuned small foundation models. Our experiments show that BM25 remains a competitive baseline compared to many zero-shot dense models in both languages. We also observe that while proprietary models outperform open alternatives in the zero-shot setting, they can be matched or surpassed by fine-tuning small language-specific models. Our dataset and evaluation code are publicly available.
\end{abstract}

\section{Introduction}
Open access to legal information is considered a fundamental right according to the Charter of Fundamental Rights in the European Union \citep{rights2024eu}. Effective retrieval models are an essential component to ensuring this right, as they allow laypeople and legal professionals to efficiently search through vast amounts of legal information. In countries like Belgium, where laws are available in multiple languages (e.g. French and Dutch), the need for high-performance legal retrieval models becomes even more crucial, as they require equal accessibility to relevant legal material regardless of the language in use.

The retrieval task \cite{thakur2beir} has experienced a significant boost due to the recent advances in textual embeddings, which rely on extensively pre-trained large language models (LLMs; \citealp{zhao2024dense}). These models can encode text into vector representations which perform very well across a broad range of tasks \cite{muennighoff2023mteb}, including classification \cite{maas2011learning,saravia2018carer,o2021wish} and clustering \cite{aggarwal2012survey,geigle2021tweac}.
Open-source models like E5 \cite{wang2022text,wang2023improving,wang2024multilingual} and BGE-M3 \cite{bgem3}, along with private models from \citet{VoyageAI2024} and \citet{OpenAI2024} have shown remarkable results in zero-shot retrieval, across multiple languages, and different domains, including various legal benchmarks \cite{muennighoff2023mteb}. These developments offer great opportunities to improve accessibility in multilingual legal jurisdictions.

Belgium invests significant resources to consolidate\footnote{\url{https://www.ejustice.just.fgov.be/cgi_loi/contenu.pl?language=nl&view_numac=2019050815nl}} its laws in both French and Dutch, which is done by the manual labor of qualified legal professionals. This results in a highly valuable resource for research in multilingual legal retrieval models. Building on this resource, and the Belgian Statutory Article Retrieval Dataset (BSARD; \citealp{louis2022statutory}) in French, we introduce the Bilingual Belgian Statutory Article Retrieval Dataset (bBSARD), which we curated by scraping parallel Dutch and French articles, and translating the BSARD questions into Dutch. Using bBSARD, we conducted extensive benchmarking  of retrieval models available for Dutch and French, both in zero-shot and fine-tuned scenarios. 

In addition to a parallel bilingual legal corpus, bBSARD offers a much-needed retrieval benchmark for the Dutch language, allowing for more accurate and reliable evaluation of Dutch retrieval models. bBSARD dataset and evaluation code are available on the HuggingFace hub\footnote{\url{https://huggingface.co/datasets/clips/bBSARD}} (under the \texttt{cc-by-nc-sa-4.0} license), and our GitHub repository\footnote{\url{https://github.com/nerses28/bBSARD}} (under the MIT license), respectively.

\section{Related Work}
In the last few years, the field of legal NLP has gained increased interest, leading to the development of a growing number of datasets for research. In this section, we focus specifically on datasets that address the task of legal retrieval grounded in legal provisions, including documents, statutory articles, and cases.

CAIL2018 \cite{xiao2018cail2018,zhong2018overview} is a dataset  designed for legal judgment prediction in Chinese, released as part of the Chinese AI and Law Challenge\footnote{\url{http://cail.cipsc.org.cn}}. It contains over 2.68 million Chinese criminal cases, linked to 183 law articles and 202 charges. One of the subtasks from this challenge involved predicting relevant law articles based on the factual descriptions of specific cases. Following this, the CAIL2019-SCM dataset \cite{xiao2019cail2019} focuses on similar case matching with 8,964 case triplets (in which two cases are similar) sourced from the Supreme People’s Court of China.

\citet{zhong2020jec} released JEC-QA, a question answering dataset based on the Chinese bar exam.  The dataset contains 26,365 multiple-choice questions, along with 3,382 Chinese legal provisions.

The AILA competitions \cite{bhattacharya2019fire,bhattacharya2020fire} introduced datasets for precedent and statute retrieval from Indian law, with content in English. For each year, around 50 queries were linked to relevant documents in retrieval corpora containing 197 statutes and around 3,000 prior cases.

Similarly, COLIEE \cite{rabelo2021coliee,rabelo2022overview,kim2022coliee,goebel2024overview} competitions include the task of statute article retrieval from provided datasets. For each year, the datasets contain around 100 test questions from the Japanese legal bar exams, labeled with relevant articles from the Japanese Civil Code, translated into English. The provided training sets include up to 1000 question-article pairs.

\citet{chen2023equals} introduced EQUALS, a dataset containing 6,914 question-article-answer triplets, with a corresponding retrieval corpus of 3,081 Chinese law articles. The question-answer pairs were collected from a free Chinese legal advice forum, then revised and further annotated by senior law students. Similarly, STARD \cite{su2024stard} introduced 1,543 queries from the general public, with a retrieval corpus of 55,348 Chinese statutory articles.

GerLayQA \cite{buttner2024answering} consists of around 21,000 legal questions from laymen paired with answers from legal professionals and grounded in paragraphs from German law books.

Most related to our work is BSARD \cite{louis2022statutory}; a statutory article retrieval dataset which contains over 1,100 legal questions from Belgian citizens and around 22,600 Belgian law articles as the retrieval corpus. LLeQA \cite{louis2024interpretable} complements BSARD with answers from legal experts, along with an additional 760 legal questions and 5,308 statutory articles. While LLeQA is a more extensive resource than BSARD, the latter is available under less restrictive terms\footnote{\url{https://huggingface.co/datasets/maastrichtlawtech/bsard}} and does not require a separate user agreement\footnote{\url{https://huggingface.co/datasets/maastrichtlawtech/lleqa}}. 

The resources presented above support the training, evaluation, and benchmarking of retrieval models across different legal domains and languages, highlighting the need for tailored approaches in each jurisdiction. Contributing to the growing field of legal NLP, we introduce bBSARD, a bilingual dataset built on BSARD which offers parallel Belgian law articles in both French and Dutch, along with legal questions translated from French to Dutch. In addition to providing a reliable benchmark for the retrieval task in Dutch, bBSARD aims to address challenges of legal retrieval in multilingual jurisdictions.

\section{Dataset}
As mentioned, we base our work on the BSARD dataset \cite{louis2022statutory}, extending it to the Dutch language by adding the corresponding articles and questions. We discuss the procedure in the following sections.

\begin{figure*}[ht]
    \centering
    \includegraphics[width=\textwidth]{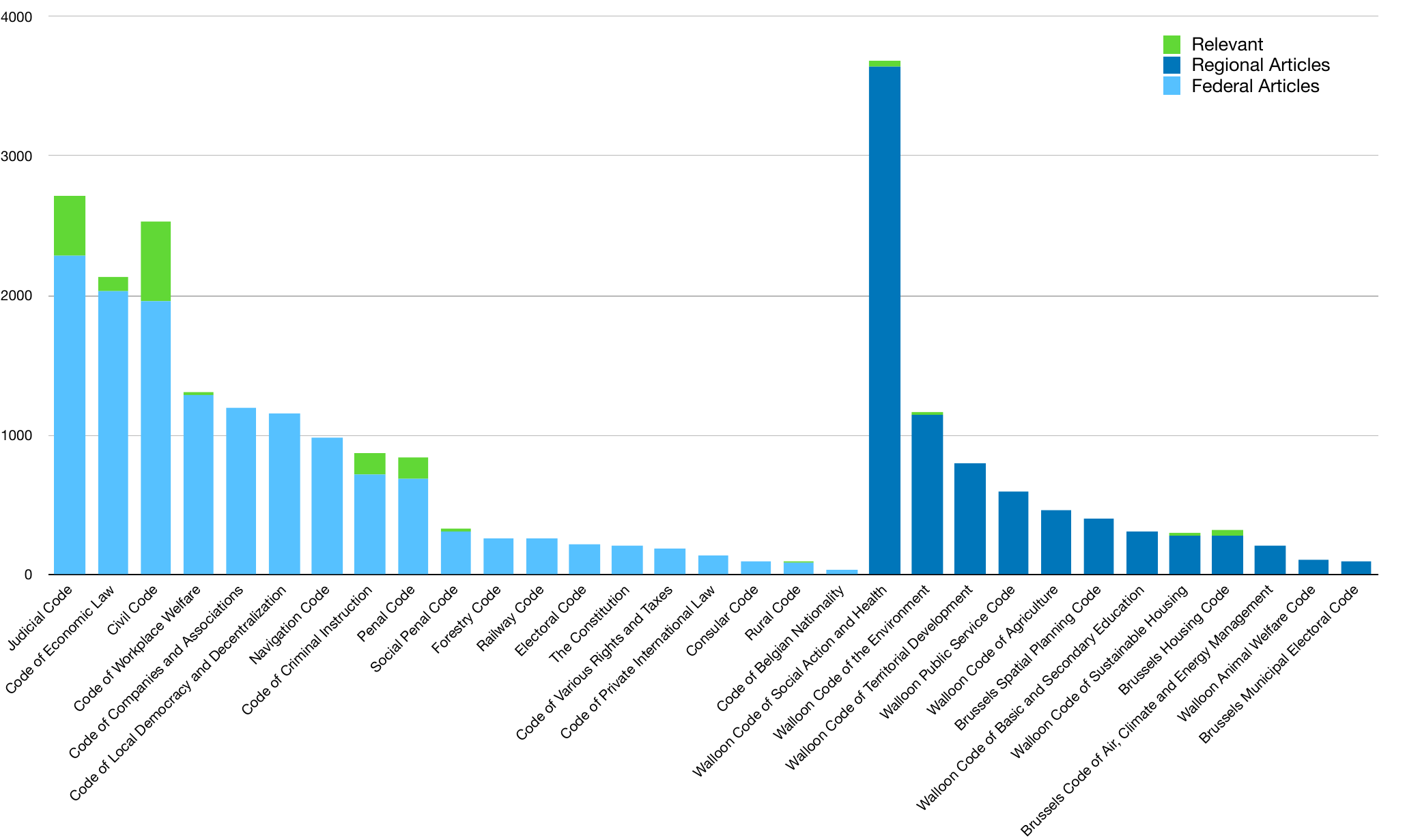}
    \caption{Distribution of different codes in the bBSARD article corpus. 'Relevant' articles (green) are the ones cited in the question set. Light and dark blue columns correspond to the Federal and Regional codes, respectively. }
    \label{fig:distrib}
\end{figure*}  

\begin{figure*}[h]
    \centering
    \includegraphics[width=\textwidth]{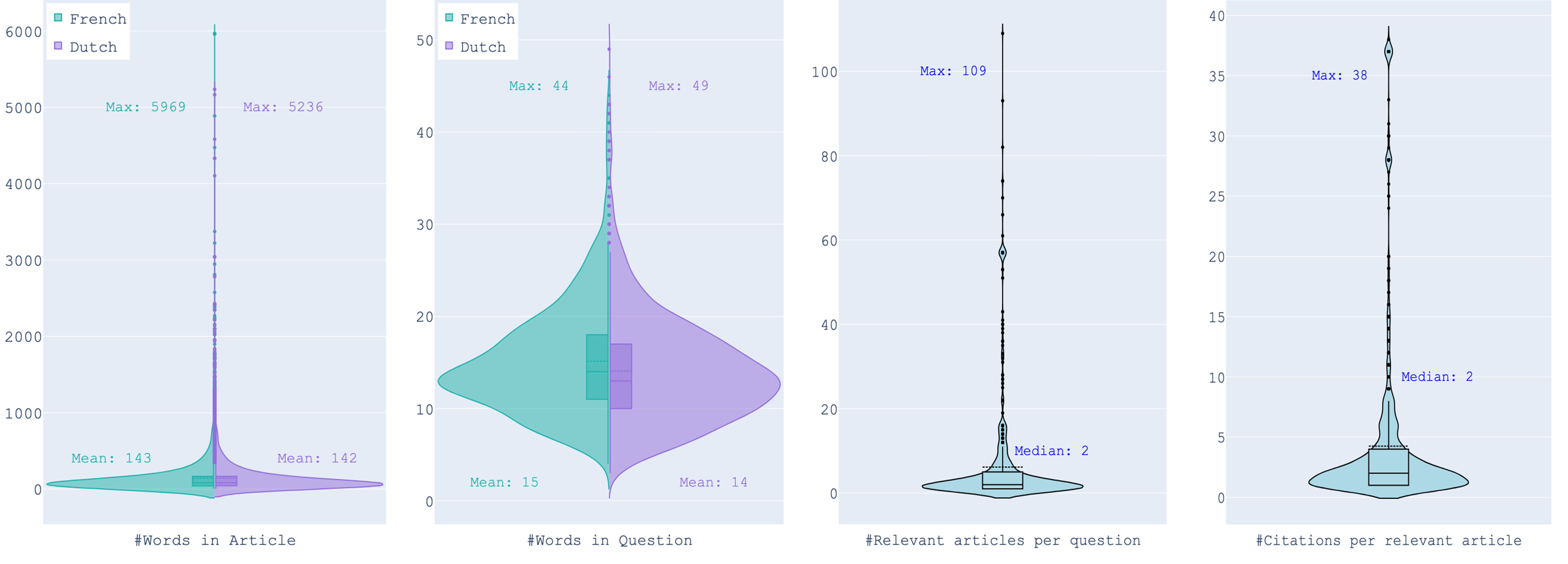}
    \caption{Basic statistics of bBSARD. From the left: Number of words in the articles (French and Dutch), Number of words in the questions (French and Dutch), number of relevant articles per question, and number of citations per relevant article. }
    \label{fig:violins}
\end{figure*}  

\subsection{Legislation in Dutch}
To get the BSARD legislation and articles in Dutch, we leverage \textbf{Justel}\footnote{\url{https://www.ejustice.just.fgov.be/cgi_loi/welcome.pl?language=nl}}, the multilingual database maintained by the Belgian Federal Government that provides online access to most Belgian legislation in French, Dutch and (often) German. Since there are no public APIs, we scrape the appropriate French and Dutch pages (52 pages for each language), according to the BSARD corpus. Considering the continuous changes and updates, and the fact that BSARD was curated in May 2021 \cite{louis2022statutory}, we make sure that the Dutch and French articles come from the same legislative version, by manually controlling their enforcement dates.

In the end, we manage to retrieve and align 22,417 out of 22,633 articles (99\%) in both languages (see Appendix \ref{sec:appendix_a} for the alignment process.). The missing 216 articles mostly belong to the Walloon Code of Environment-Decrees (126 articles absent from the Dutch page), and the Military Penal Code (66 articles absent from the database). Fortunately, these missing articles contribute only marginally to the relevant subset (only 1 missing article is cited in a multi-referenced question). Table \ref{tab:distrib_compr} in Appendix \ref{sec:appendix_a} summarizes the differences between the original and bilingual datasets. 

Figure \ref{fig:distrib} shows how different codes contribute to the complete and relevant set of articles. The majority of relevant articles (i.e. annotated as necessary to answer questions, colored light green in the chart) come from four Federal codes: Judicial, Civil, Penal, and Criminal Instruction.

\subsection{Questions in Dutch}
\label{sec:dutch_trans}
BSARD contains 1,108 questions (split as 886/222 for the train/test sets), each labeled by experts with the IDs of the corresponding relevant law articles from the corpus. These questions have been curated in partnership with Droits Quotidiens\footnote{\url{https://droitsquotidiens.be/}}, from emails sent by Belgian citizens to this organization, asking for advice on legal issues. They cover a wide range of topics, with around 85\% of them being either about family, housing, money, or justice, while the remaining 15\% concern either social security, foreigners, or work \cite{louis2022statutory}.

To produce these questions in Dutch, we opt for automatic translation followed by human inspection. We first prompt OpenAI's GPT-4o with the original French question, as well as the relevant articles (to provide context), and ask for the Dutch translation (The full prompt can be found in Appendix \ref{sec:appendix_c}). To increase translation fidelity, we set the temperature to 0 \cite{peng-etal-2023-towards}.  We then asked a native speaker to examine a random sample of 100 translated questions, and annotate them for potential issues. The results showed (legally) inaccurate choice of words in 2\%, and minor semantic/grammatical/lexical issues (e.g. translation being too literal) in 6\% of the studied samples. 

Figure \ref{fig:violins} shows basic statistical features of the bBSARD dataset. The French and Dutch articles have an average length of 143 and 142 words, respectively, while for the questions these numbers stand at 15 and 14 words. Regarding the question-article mapping, 1,611 distinct articles (out of 22,417) contribute to the relevant subset, and 75\% of questions have fewer than five references, with a median value of two.

\section{Experimental Setup}
This section describes our experimental setup used to benchmark the retrieval performance of a selection of models on bBSARD. We mostly reuse the codebase from BSARD \cite{louis2022statutory}, making modifications where necessary to accommodate the retrieval models to the specific requirements of our experiments. Below we describe the models, data processing steps, and evaluation metrics used in our experiments.

\subsection{Models}
We select a diverse range of models in three different categories/settings: lexical, zero-shot, and fine-tuned. 

\subsubsection{Lexical models}
Lexical approaches for retrieval rely on keyword matching and utilize various word (or token) weighting schemes and algorithms to determine the relevance of documents for a given query. The most popular algorithms are TF-IDF (Term Frequency-Inverse Document Frequency; \citealp{sparck1972statistical,salton1973specification}) and BM25 (Best Match 25; \citealp{robertson-1994-okapi}). Despite the lexical gap issues, where the vocabulary used in queries can differ from that in relevant documents,  BM25 remains a robust baseline for many retrieval tasks. Remarkably BM25 was outperformed only recently by E5 \cite{wang2022text} on the BEIR retrieval benchmark \cite{thakur2beir} in a setup that does not utilize any labeled data. In our experiments, we evaluate both TF-IDF and BM25.

\begin{figure*}[ht]
    \centering
    \includegraphics[width=.8\textwidth]{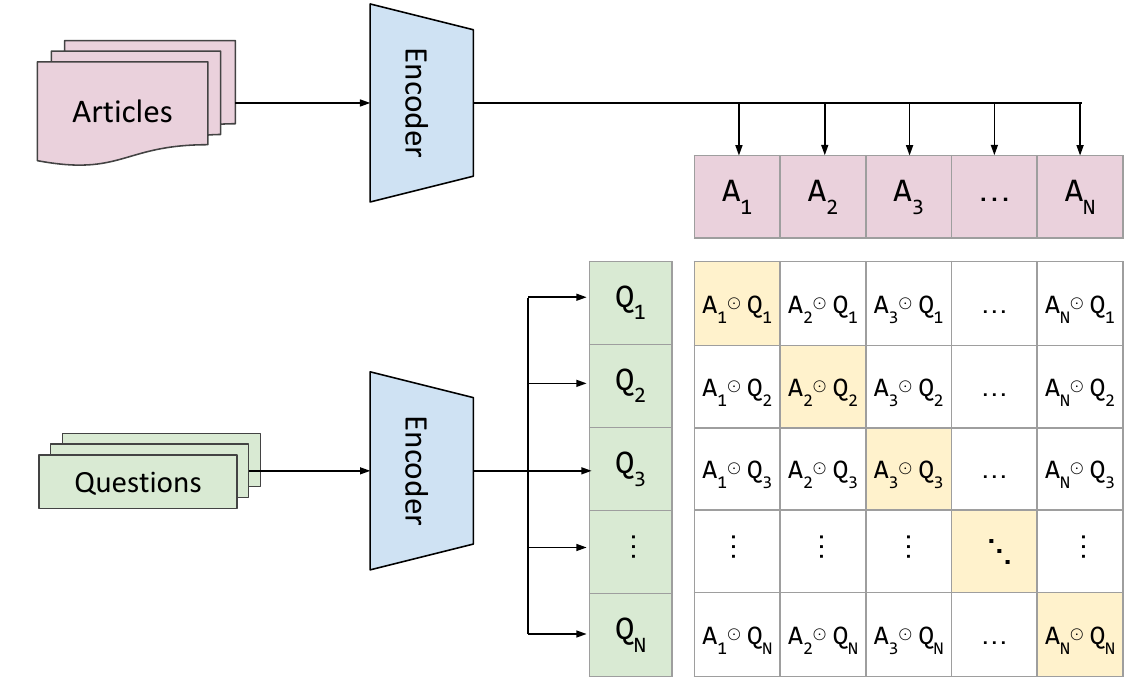}
    \caption{Standard Siamese Bi-Encoder setting with in-batch negatives, which we use for fine-tuning. Articles and Questions are encoded separately with the same model into vectors. For each question $Q_{i}$, the relevant article $A_{i}$ is the positive sample, while all other articles in the batch are used as negatives. $\odot$ represents the cosine similarity operator.}
    \label{fig:arch}
\end{figure*}

\subsubsection{Zero-shot models}
Recently, LLMs achieved impressive results on various retrieval tasks \cite{zhao2024dense}. For the zero-shot setting, we select the following multilingual retrieval models, from both open and proprietary categories: mContriever\footnote{\url{https://huggingface.co/facebook/mcontriever-msmarco}} \cite{izacardunsupervised}, LaBSE \cite{feng2022language},   mE5 \cite{wang2024multilingual}, E5$_{mistral-7b}$ \cite{wang2023improving}, BGE-M3 \cite{bgem3}, DPR-XM \cite{louis2024colbert}, BGE-Multilingual-Gemma2 \cite{li2024making}, jina-embeddings-v3 \cite{sturua2024jina}, mGTE \cite{zhang2024mgte}, voyage-3 \cite{VoyageAI2024}, text-embedding-3-large \cite{OpenAI2024}. For models with a maximum input length of 512 tokens (except LaBSE), we divide the text into overlapping chunks of 200 tokens with an overlap of 20 tokens between neighboring chunks to mitigate the input length limitations. We do not impose any limits on the input length for other models, allowing them to handle truncation if necessary.

In addition, we experiment with context-independent word embeddings, using
word2vec \cite{mikolov2013distributed,Mikolov2013EfficientEO} for Dutch \cite{tulkens2016evaluating} and French \cite{fauconnier_2015}, as well as fastText \cite{bojanowski2017enriching} for both Dutch and French \cite{grave2018learning}. To construct embeddings of text chunks from these models, we apply mean-pooling to the word embeddings, with the exception of LaBSE, which uses the [CLS] token representation.  In all cases, cosine similarity is employed to score similarity between the embeddings. 

The evaluation is conducted on a single GPU with 48GB of RAM for E5$_{mistral-7b}$ and BGE-Multilingual-Gemma2. For other models, we use a single GPU with 8GB of RAM. Each experiment takes between five minutes for smaller models and up to 30 minutes for larger models.

\subsubsection{Fine-tuned models}
Foundation models can achieve competitive results compared to zero-shot retrieval models when fine-tuned on domain-specific data. In our evaluations, we select  RobBERT-2023 \cite{delobelle2024robbert} and Tik-to-Tok \cite{remy2023tik} for Dutch, and CamemBERT \cite{martin2020camembert} and Flaubert \cite{le2020flaubert} for French. We also include XLM-Roberta to examine the potential advantage of language-specific models over the generic multilingual ones. 

We primarily follow the experimental setup of BSARD and fine-tune the models in a Siamese setting \cite{Reimers2019SentenceBERTSE}, which encodes the query and document via the same model (Figure \ref{fig:arch}). We optimize the contrastive loss with a temperature of 0.05 and in-batch negatives \cite{henderson2017efficient,karpukhin2020dense} with a batch size of 22. The optimization is performed using AdamW \cite{Loshchilov2017DecoupledWD} with a learning rate of 2e-5, $\beta_1 = 0.9$, $\beta_2 = 0.999$, and weight decay of 0.01. The learning rate undergoes a warm-up over the first 500 steps, followed by linear decay. Following \citet{louis2022statutory}, training is performed for 100 epochs, which takes 4.5-5.5 hours (depending on model size) on a single GPU with 24GB of RAM. Finally, we employ cosine similarity to score the embeddings. 

\begin{table*}[hp]
\centering
\small % Adjust font size to fit table
\begin{tabular}{llcccccccc}
\hline
\textbf{T} & \textbf{Model} & \textbf{Size} &  \textbf{R@100} & \textbf{R@200} & \textbf{R@500} & \textbf{MAP@100} & \textbf{MRR@100} & \textbf{nDCG@10} & \textbf{nDCG@100} \\
\hline
 & TF-IDF & - & 39.21 & 46.38 & 52.76 & 8.53 & 14.25 & 12.38 & 16.74 \\
 & BM25 & - & 40.19 & 47.95 & 54.57 & 16.07 & 22.63 & 20.07 & 23.57 \\
\hline
 & word2vec & - & 41.06 & 51.02 & 58.94 & 8.28 & 15.27 & 11.66 & 17.05 \\
 & fastText & - & 31.47 & 38.26 & 49.08 & 7.27 & 12.67 & 8.89 & 13.86\\
  & mE5$_{small}$ & 118M & 45.43 & 52.25 & 61.10 & 13.42 & 21.79 & 17.67 & 22.79 \\
  & mContriever & 178M & 47.92 & 58.38 & 68.32 & 11.38 & 20.15 & 14.83 & 21.82 \\
   & DPR-XM & 277M & 40.44 & 46.12 & 53.16 & 13.57 & 21.78 & 16.40 & 21.79\\
  & mE5$_{base}$ & 278M & 50.14 & 57.68 & 65.30 & 16.47 & 25.64 & 20.81 & 26.49 \\
   & mGTE & 305M & 52.78 & 61.97 & 73.09 & 15.86 & 24.92 & 20.08 & 26.80 \\
   & LaBSE  & 471M & 20.51 & 28.42 & 42.18 & 2.34 & 6.60 & 3.50 & 7.18  \\
   & mE5$_{large}$ & 560M & 58.35 & 65.83 & 70.83 & 21.88 & 34.28 & 28.47 & 33.51 \\
    & mE5$_{large-instruct}$ & 560M & 59.48 & 66.80 & 75.21 & 18.66 & 29.93 & 24.84 & 31.33 \\

 & BGE-M3 & 568M & 61.12 & 67.20 & 77.56 & 18.31 & 30.40 & 24.04 & 31.21 \\
 & jina-embeddings-v3 & 572M &  60.70 & 67.92 & 77.37 & 18.59 & 31.21 & 24.70 & 31.58 \\
   & E5$_{mistral-7b}$ & 7B  & 68.35 & 73.91 & 82.82 & 30.24 & 43.26 & 37.70 & 43.02 \\
  & BGE-Mult.-Gemma2 & 9B & 69.94 & 76.23 & 81.28 & 25.07 & 37.66 & 30.95 & 39.11 \\
 & voyage-3 & - & 73.08 & 79.37 & 85.67 & 32.81 & 46.38 & 40.06 & 46.21  \\
 & embedding-3-large & - & \textbf{75.70} & \textbf{80.22} & \textbf{88.24} & 29.73 & 42.99 & 36.83 & 44.40 \\
\hline
\checkmark &Tik-to-Tok$_{base}$ & 116M & 73.90 & 79.02 & 83.29 & 39.24 & 45.69 & 42.75 & 49.90 \\
\checkmark & RobBERT-2023$_{base}$ & 125M & 75.08 & 79.33 & 83.40 & \textbf{40.51} & \textbf{47.68} & \textbf{44.76} & \textbf{51.36} \\
\checkmark &  XLM-Roberta$_{base}$ & 279M & 62.06 & 68.26 & 75.40 & 26.61 & 32.00 & 30.65 & 37.10 \\
\hline
\end{tabular}
\newline
\caption{Retrieval performance of different models on the Dutch subset of bBSARD (test set). Evaluations are zero-shot for the dense models, except for the last 3 models (check-marked) which are fine-tuned. }
\label{tab:nl_performance}
\vspace{1em}

\begin{tabular}{llcccccccc}
\hline
\textbf{T} & \textbf{Model} & \textbf{Size} & \textbf{R@100} & \textbf{R@200} & \textbf{R@500} & \textbf{MAP@100} & \textbf{MRR@100} & \textbf{nDCG@10} & \textbf{nDCG@100} \\
\hline
 & TF-IDF & -  & 41.69 & 51.05 & 60.22 & 8.74 & 12.85 & 11.34 & 17.45 \\
 & BM25 & - & 51.81 & 56.95 & 65.51 & 17.02 & 26.02 & 21.54 & 27.52 \\
\hline
 & word2vec & -  & 49.93 & 62.29 & 71.11 & 13.45 & 21.45 & 17.32 & 23.62 \\
 & fastText & -  & 24.84 & 32.36 & 43.88 & 5.05 & 10.03 & 7.40 & 10.65 \\
  & mE5$_{small}$  & 118M & 46.26 & 51.74 & 59.25 & 13.67 & 23.48 & 18.49 & 23.03 \\
  & mContriever &  178M  & 46.01 & 56.62 & 68.42 & 12.94 & 21.56 & 17.59 & 22.94 \\
& DPR-XM & 277M & 40.91 & 47.34 & 55.13 & 10.83 & 19.31 & 14.31 & 19.74\\
  & mE5$_{base}$ & 278M & 47.62 & 56.60 & 63.60 & 16.76 & 26.25 & 21.90 & 26.28 \\
  & mGTE & 305M & 57.54 & 66.57 & 77.02 & 19.40 & 30.14 & 24.13 & 31.02 \\
  & LaBSE  & 471M & 21.62 &  32.86 & 46.66 & 2.74 &   7.00 & 4.17 & 7.67 \\
   & mE5$_{large}$ & 560M & 55.30 & 62.83 & 69.85 & 21.54 & 34.27 & 28.06 & 32.68 \\
    & mE5$_{large-instruct}$  & 560M & 60.99 & 68.34 & 76.75 & 19.77 & 32.60 & 26.52 & 32.44 \\
 & BGE-M3 & 568M & 60.76 & 69.02 & 79.81 & 19.40 & 31.38 & 25.38 & 32.08 \\
 & jina-embeddings-v3 & 572M & 64.05 & 71.67 & 78.76 & 20.51 & 34.52 & 27.09 & 34.19 \\
   & E5$_{mistral-7b}$ &7B & 69.41 & 74.53 & 84.06 & 27.43 & 40.22 & 34.82 & 41.07 \\
 & BGE-Mult.-Gemma2 & 9B & 71.44 & 77.81 & 83.73 & 30.06 & 43.72 & 36.36 & 43.46 \\
 & voyage-3 & -  & 77.71 & \textbf{82.68} & \textbf{88.76} & 38.78 & \textbf{54.60} & 45.96 & 52.51 \\
 & embedding-3-large & - & 75.47 & 80.70 & 87.58 & 33.72 & 46.51 & 40.54 & 47.33\\
\hline
\checkmark & CamemBERT$_{base}$ & 111M & 77.10 & 80.63 & 86.37 & 39.08 & 46.99 & 44.25 & 50.95 \\
\checkmark & FlauBERT$_{base}$ & 138M & \textbf{78.15} & 81.59 & 85.84 & \textbf{42.11} & 49.82 & \textbf{46.69} & \textbf{53.48} \\
\checkmark &  XLM-Roberta$_{base}$ & 279M &  63.31 & 70.70 & 77.76 & 30.57 & 37.84 & 34.90 & 40.82 \\

\hline
\end{tabular}
\newline
\caption{Retrieval performance of different models on the French subset of bBSARD (test set). Evaluations are zero-shot for the dense models, except for the last 3 models (check-marked) which are fine-tuned. }
\label{tab:fr_performance}
\end{table*}

\subsection{Metrics}
To assess the performance of our models, we employ standard retrieval metrics: macro-averaged recall@k (R@k), mean average precision@k (MAP@k), mean reciprocal rank@k (MRR@k), and normalized discounted cumulative gain@k (nDCG@k).

\section{Results and Discussion}
In this section, we present the performance of various retrieval models evaluated on the Dutch and French subsets of the bBSARD dataset (see Tables \ref{tab:nl_performance} and \ref{tab:fr_performance}). In addition, we directly compare model effectiveness between two languages leveraging the parallel nature of the dataset.
\subsection{Dutch Subset}
As Table \ref{tab:nl_performance} shows, BM25 proves to be a strong baseline for the Dutch subset, with zero-shot dense models only fully outperforming it starting from 300 million parameters.

In the zero-shot setting, we observe a consistent improvement in performance as the model size grows, with the exception of LaBSE, which shows relatively lower results. The small-sized models (below 200M parameters), mE5$_{small}$ and mContriever, outperform the context-independent models (i.e. word2vec and fastText), and while mContriever achieves higher recall (R@100, R@200, R@500), mE5$_{small}$ is better across all other metrics. mE5$_{small}$ even outperforms the larger DPR-XM model in almost all metrics. 

In the next zero-shot category (around 300M parameters), mGTE significantly outperforms mE5$_{base}$ in recall (R@100, R@200, R@500), while doing marginally worse across other metrics. For models up to 1 billion parameters, BGE-M3 and jina-embeddings-v3 show comparable results and are the best performers in recall (R@100, R@200, R@500), but E5$_{large}$ demonstrates superior performance in MAP@100, MRR@100, and nDCG (@10, @100). Finally, the largest open models, E5$_{mistral-7b}$ and BGE-Multilingual-Gemma2, outperform all other open models by a large margin. However, they lag behind proprietary models, voyage-3 and embedding-3-large, which are the best performers for the zero-shot setup. 

As the lower section of the table shows, the high performance of proprietary models can be matched or topped by fine-tuning small models. In particular, fine-tuned RobBERT-2023$_{base}$ outperforms these models in MAP, MRR and nDCG metrics. Additionally, language-specific models demonstrate a significant advantage over the multilingual XLM-Roberta.

\subsection{French Subset}
Table \ref{tab:fr_performance} shows the results for the French subset of bBSARD. We observe trends similar to Dutch, with BM25 remaining competitive with the zero-shot dense models up to 300 million parameters.

Similarly, we observe a steady increase in performance in the zero-shot setup as the average model size increases, with the exception of LaBSE. Interestingly, the context-independent model word2vec outperforms not only the sub-200M models mE5$_{small}$ and mContriever, but also the larger DPR-XM model, while beating mE5$_{base}$ in recall. In the 300M-parameter category, mGTE outperforms the larger mE5$_{large}$ model in recall (R@100, R@200, R@500), and competes with BGE-M3 in MAP@100, MRR@100, and nDCG (@10, @100). Among the models up to 1 billion parameters, jina-embeddings-v3 achieves the highest performance in recall (R@100, R@200), MRR@100, and nDCG@100, while BGE-M3 performs better in recall@500, and E5$_{large}$ demonstrates the best results in nDCG@10. The largest open models, E5$_{mistral-7b}$ and BGE-Multilingual-Gemma2, show superior performance over other open options. However, the proprietary models, voyage-3 and embedding-3-large, outperform them by a large margin, with voyage-3 showing the best overall performance. 
Finally, we see the competitive performance of small fine-tuned models, with FlauBERT$_{base}$ beating voyage-3 in 4 out of 7 metrics.

\subsection{Cross-Language Comparison}

As bBSARD is a parallel dataset, we can directly compare Tables \ref{tab:nl_performance} and \ref{tab:fr_performance} to gain deeper insights into performance discrepancies between the French and Dutch models. 

On average, models show a higher performance on the French subset compared to the Dutch subset (see Table \ref{tab:fr_nl_diff} in Appendix \ref{sec:appendix_b}). This is perhaps most notable in BM25 and word2vec, where French models outperform their Dutch counterparts by more than 10 recall points (the clear outlier is fastText, which performs significantly better on the Dutch subset.) In addition, mGTE, jina-embeddings-v3, and voyage-3 do significantly better on the French subset than the Dutch. Other models gain 2-3 additional recall points in Dutch and perform comparably well across other metrics for both languages, with the exception of E5$_{mistral-7b}$ and DPR-XM. These models show slightly lower recall (R@100, R@200, R@500) in Dutch but achieve higher scores in other metrics. Finally, the best fine-tuned performer in French, FlauBERT$_{base}$, outperforms the top performer in Dutch, RobBERT-2023$_{base}$, and XLM-Roberta gains 3-5 points higher results when trained and evaluated on French. 

In addition to potential translation issues (see \ref{sec:dutch_trans}) which particularly affect lexical models, one intuitive hypothesis on the origin of this advantage concerns the significant difference in data availability between the two languages. For example, while the original RobBERT model was pre-trained on a 39 GB corpus \cite{delobelle-etal-2020-robbert}, CamemBERT and FlauBERT used 138 GB and 71 GB of data, respectively\footnote{282 GB and 270 GB before filtering/cleaning.} \cite{le2020flaubert}. However, further analysis is required to determine the significance of this factor, as well as other contributing parameters.       

\section{Conclusions and Future Work}
In this paper we presented bBSARD, the bilingual version of the BSARD dataset \cite{louis2022statutory}. To curate bBSARD, we scraped parallel Dutch and French articles from the online Justel database and translated the BSARD questions into Dutch. In addition to a parallel bilingual legal corpus, bBSARD offers a much-needed retrieval benchmark for the Dutch language, allowing for more accurate and reliable evaluation of Dutch retrieval models. 

Based on our dataset, we conducted extensive benchmarking of the retrieval task (ranking passages by their relevance to a given query) for Dutch and French, both in 
zero-shot and fine-tuned scenarios. These experiments confirm the status of simple lexical methods like BM25 as strong baselines, the superiority of closed-source commercial models like Voyage and OpenAI in zero-shot setting, and the possibility of outperforming them via fine-tuning small language-specific models like RobBERT and FlauBERT. We also observed an overall advantage for French compared to Dutch, in both zero-shot and fine-tuning scenarios. 

We hope that our work encourages and facilitates the development of better Dutch retrieval models in the legal domain, which are an essential part of popular LLM-based methods like RAG. In the future, we would like to first improve bBSARD's quality by manually checking/correcting all translated questions, and then expand our work beyond the legal domain by curating and providing a more comprehensive benchmark for the retrieval task in Dutch. Another interesting research avenue considers the cross-lingual training potentials offered by a bilingual parallel dataset. In our experiments, we observed that XLM-Roberta performs better in Dutch when finetuned for 50+50 epochs on French+Dutch data, compared to 100 epochs on the Dutch subset. This suggests the possibility of leveraging the bilingual structure for additional gains in performance, specially for the lower-resource language, but to examine and explore its real significance more experiments need to be conducted. 

\subsection*{Limitations}
We primarily inherit limitations from BSARD, as this dataset serves as the foundation for our work. The retrieval corpus is limited to the 32 Belgian codes from federal (Belgian) and Walloon law. As a result, bBSARD does not cover the whole of Belgian law, particularly omitting codes specific to Flanders. In addition, these limitations make the retrieval process incomplete as a part of relevant articles might be missing.  Since we scraped the Belgian articles from around May 2021, bBSARD does not contain the updated version of the Belgian law. 

Given these limitations, bBSARD is not intended for obtaining any comprehensive legal information or advise.  Its primary purpose is to benchmark retrieval models and gain insights into the current state of the art. In accordance with the BSARD license (\texttt{cc-by-nc-sa-4.0}), we release our dataset under the same terms.

\section*{Acknowledgments}
This research received funding from the Flemish Government under the “Onderzoeksprogramma Artificiële Intelligentie (AI) Vlaanderen” programme.
We would like to thank Franci Haest for contributing to the data collection and to Luna De Bruyne for checking the quality of the translations. We also acknowledge the use of the GPT-4o model for assisting with error checking and proofreading of this paper.

\bibliography{custom}
\appendix

\section{Appendix: Scraping and Aligning the Articles}
\label{sec:appendix_a}
Table \ref{tab:distrib_compr} shows a detailed summary of codes scraped from the \textbf{Justel} portal for bBSARD, compared to the original BSARD dataset. For alignment, we first leverage the article names/codes (e.g. Art. 14bis), and then use an automatic pipeline (consisting of a length comparison filter followed by ChatGPT queries) to spot the absent, misaligned, or non-aligned articles, which we then add and/or align manually. The alignment issues are mainly due to rare discrepancies in the way articles are registered in French and Dutch pages, or between French pages and BSARD dataset (for example `Art. 14.2' vs. `Art. 14/2').

\begin{table*}[h]
\centering
\small % Adjust font size to fit table
\begin{tabular}{|l|l|cc|cc|}
\hline
&& \multicolumn{2}{c|}{\textbf{BSARD}} & \multicolumn{2}{c|}{\textbf{bBSARD}} \\
\textbf{Authority} &\textbf{Code} & \#Articles & \#Relevant & \#Articles & \#Relevant \\
\hline
Federal & Judicial Code & 2285 & 429 & 2283 & 429 \\
&Code of Economic Law & 2032 & 98 &  2032 & 98 \\
&Civil Code & 1961 & 568 & 1961 & 568 \\
&Code of Workplace Welfare & 1287 & 25 & 1287 & 25 \\
&Code of Companies and Associations & 1194 & 0 & 1193 & 0 \\
&Code of Local Democracy and Decentralization & 1159 & 3 & 1158 & 3 \\
&Navigation Code & 977 & 0& 977 & 0 \\
&Code of Criminal Instruction & 719 & 155 & 719 & 155 \\
&Penal Code & 689 & 154 & 689 & 154 \\ 
&Social Penal Code & 307 & 23 & 307 & 23 \\
&Forestry Code & 261& 0& 261& 0 \\
&Railway Code & 260 & 0 & 260 & 0 \\ 
&Electoral Code & 218 & 0 & 217 & 0 \\
&The Constitution & 208 & 5 & 208 & 5 \\
&Code of Various Rights and Taxes & 191 & 0 & 189 & 0 \\
&Code of Private International Law & 135 & 4 & 134 & 4 \\
&Consular Code & 100 & 0& 100 & 0 \\
&Rural Code & 87 & 12 & 87 & 12 \\
&Military Penal Code & 66 & 1 & 0 & 0 \\
&Code of Belgian Nationality & 31 & 8 & 31 & 8 \\
\hline
Regional &Walloon Code of Social Action and Health & 3650 & 40 & 3643 & 40 \\
&Walloon Code of the Environment & 1270 & 22 & 1143 & 22 \\
&Walloon Code of Territorial Development & 796 & 0 & 795 & 0 \\ 
&Walloon Public Service Code & 597 & 0 & 597 & 0 \\
&Walloon Code of Agriculture & 461 & 0 & 461 & 0 \\
&Brussels Spatial Planning Code & 401 & 1 & 401 & 1 \\
&Walloon Code of Basic and Secondary Education & 310 & 0 & 310 & 0 \\
&Walloon Code of Sustainable Housing & 286 & 20 & 279 & 20 \\
&Brussels Housing Code & 279 &44 & 279 &44 \\ 
&Brussels Code of Air, Climate and Energy Management & 208 & 0 & 208 & 0 \\
&Walloon Animal Welfare Code & 108 & 0 & 108 & 0 \\
&Brussels Municipal Electoral Code & 100 & 0 & 100 & 0 \\
\hline
&Total & 22633 & 1612 & 22417 & 1611\\
\hline
\end{tabular}
\newline
\caption{Distribution of codes in BSARD and bBSARD (this work). "Relevant" articles are meant with respect to the question set.}
\label{tab:distrib_compr}
\end{table*}

\section{Appendix: Translating the Questions }
\label{sec:appendix_c}
To translate the questions into Dutch, we prompt \texttt{GPT-4o}\footnote{gpt-4o-2024-08-06} with the following instruction and context for each question (temperature = 0).

\textbf{Prompt:}\texttt{"You will be provided with a legal question and a related article from Belgian legislation. Your task is to translate the question from French to Dutch. The article serves solely as context to ensure the accuracy in legal understanding and terminology, so do not include any part of it in the translation. Return only the translation of the question without any additional information.}

\texttt{<article>: \{article\} </article>}

\texttt{<question>: \{question\} </question>}

\texttt{question translated to Dutch:"} \\

We also translate the 3 meta-fields available for each question in BSARD, i.e. \texttt{category}, \texttt{subcategory}, \texttt{extra\_description} (although they are not used in the experiments). For this, we first refer to the \texttt{\url{www.helderrecht.be}} website (the Dutch version for \texttt{\url{www.droitsquotidiens.be}}), and extract the available corresponding categories and subcategories (35\% of the total). We then use these translation pairs as examples to prompt \texttt{GPT-4o} to translate the rest of the phrases.     

\section{Appendix: Comparison of French and Dutch Results }
\label{sec:appendix_b}
Table \ref{tab:fr_nl_diff} shows the average retrieval performance for different model types on the French and Dutch subsets of bBSARD (test set).
 
\begin{table*}[hp]
\centering
\small % Adjust font size to fit table

\begin{tabular}{llcccccccc}
\hline
\textbf{T} & \textbf{Model Type} & \textbf{Lang.} & \textbf{R@100} & \textbf{R@200} & \textbf{R@500} & \textbf{MAP@100} & \textbf{MRR@100} & \textbf{nDCG@10} & \textbf{nDCG@100} \\
\hline
 & Lexical & FR  & \textbf{46.75} & \textbf{54.00} & \textbf{62.87} & \textbf{12.88} & \textbf{19.44} & \textbf{16.44} & \textbf{22.49} \\
 &  & NL & 39.70 & 47.17 & 53.67 & 12.30 & 18.44 & 16.23 & 20.16 \\
\hline
 & CI dense & FR  & \textbf{37.39} & \textbf{47.33} & \textbf{57.50} & \textbf{9.25} & \textbf{15.74} & \textbf{12.36} & \textbf{17.14} \\
 &  & NL  & 36.27 & 44.64 & 54.01 & 7.78 & 13.97 & 10.28 & 15.46 \\
 \hline
  & CD dense  & FR & \textbf{56.79} & \textbf{64.24} & \textbf{72.81} & \textbf{20.54} & \textbf{31.83} & \textbf{26.09} & \textbf{31.89} \\
  &  &  NL  & 56.00 & 63.02 & 71.51 & 19.17 & 29.79 & 24.35 & 30.52 \\
\hline
\checkmark & CD dense & FR & \textbf{72.85} & \textbf{77.64} & \textbf{83.32} & \textbf{37.25} & \textbf{44.88} & \textbf{41.95} & \textbf{48.42} \\
 &  & NL & 70.35 & 75.54 & 80.70 & 35.45 & 41.80 & 39.39 & 46.12 \\

\hline
\end{tabular}
\newline
\caption{Average retrieval performance per model type on bBSARD (test set). CI 
and CD refer to context-independent and context-dependent models, respectively. All dense models are evaluated in zero-shot setting, except for the lower section (check-marked) which are fine-tuned. }
\label{tab:fr_nl_diff}
\end{table*}

\end{document}